# The Design and Experimental Analysis of Algorithms for Temporal Reasoning

**Peter van Beek**                              VANBEEK@CS.UALBERTA.CA
**Dennis W. Manchak**                           DMANCHAK@VNET.IBM.COM
*Department of Computing Science, University of Alberta*
*Edmonton, Alberta, Canada T6G 2H1*

## Abstract

Many applications—from planning and scheduling to problems in molecular biology—rely heavily on a temporal reasoning component. In this paper, we discuss the design and empirical analysis of algorithms for a temporal reasoning system based on Allen's influential interval-based framework for representing temporal information. At the core of the system are algorithms for determining whether the temporal information is consistent, and, if so, finding one or more scenarios that are consistent with the temporal information. Two important algorithms for these tasks are a path consistency algorithm and a backtracking algorithm. For the path consistency algorithm, we develop techniques that can result in up to a ten-fold speedup over an already highly optimized implementation. For the backtracking algorithm, we develop variable and value ordering heuristics that are shown empirically to dramatically improve the performance of the algorithm. As well, we show that a previously suggested reformulation of the backtracking search problem can reduce the time and space requirements of the backtracking search. Taken together, the techniques we develop allow a temporal reasoning component to solve problems that are of practical size.

## 1. Introduction

Temporal reasoning is an essential part of many artificial intelligence tasks. It is desirable, therefore, to develop a temporal reasoning component that is useful across applications. Some applications, such as planning and scheduling, can rely heavily on a temporal reasoning component and the success of the application can depend on the efficiency of the underlying temporal reasoning component. In this paper, we discuss the design and empirical analysis of two algorithms for a temporal reasoning system based on Allen's (1983) influential interval-based framework for representing temporal information. The two algorithms, a path consistency algorithm and a backtracking algorithm, are important for two fundamental tasks: determining whether the temporal information is consistent, and, if so, finding one or more scenarios that are consistent with the temporal information.

Our stress is on designing algorithms that are robust and efficient in practice. For the path consistency algorithm, we develop techniques that can result in up to a ten-fold speedup over an already highly optimized implementation. For the backtracking algorithm, we develop variable and value ordering heuristics that are shown empirically to dramatically improve the performance of the algorithm. As well, we show that a previously suggested reformulation of the backtracking search problem (van Beek, 1992) can reduce the time and space requirements of the backtracking search. Taken together, the techniques we develop allow a temporal reasoning component to solve problems that are of realistic size. As part of the evidence to support this claim, we evaluate the techniques for improving the algorithms on a large problem that arises in molecular biology.

| Relation | Symbol | Inverse | Meaning |
|----------|--------|---------|---------|
| x before y | b | bi | 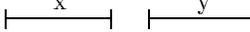 |
| x meets y | m | mi | 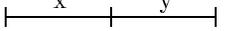 |
| x overlaps y | o | oi | 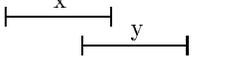 |
| x starts y | s | si | 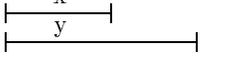 |
| x during y | d | di | 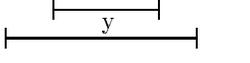 |
| x finishes y | f | fi | 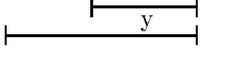 |
| x equal y | eq | eq | 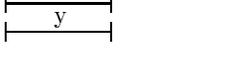 |

Figure 1: Basic relations between intervals

## 2. Representing Temporal Information

In this section, we review Allen's (1983) framework for representing relations between intervals. We then discuss the set of problems that was chosen to test the algorithms.

### 2.1 Allen's framework

There are thirteen *basic* relations that can hold between two intervals (see Figure 1; Allen, 1983; Bruce, 1972). In order to represent indefinite information, the relation between two intervals is allowed to be a disjunction of the basic relations. Sets are used to list the disjunctions. For example, the relation {m,o,s} between events A and B represents the disjunction, (A meets B) ∨ (A overlaps B) ∨ (A starts B). Let $I$ be the set of all basic relations, {b,bi,m,mi,o,oi,s,si,d,di,f,fi,eq}. Allen allows the relation between two events to be any subset of $I$.

We use a graphical notation where vertices represent events and directed edges are labeled with sets of basic relations. As a graphical convention, we never show the edges $(i, i)$, and if we show the edge $(i, j)$, we do not show the edge $(j, i)$. Any edge for which we have no explicit knowledge of the relation is labeled with $I$; by convention such edges are also not shown. We call networks with labels that are arbitrary subsets of $I$, interval algebra or IA networks.

**Example 1.** Allen and Koomen (1983) show how IA networks can be used in non-linear planning with concurrent actions. As an example of representing temporal information using IA networks, consider the following blocks-world planning problem. There are three blocks, A, B, and C. In the initial state, the three blocks are all on the table. The goal state is simply a tower of the blocks with A on B and B on C. We associate states, actions, and properties with the intervals they hold over, and we can immediately write down the following temporal information.

| *Initial Conditions* | *Goal Conditions* |
| --- | --- |
| Initial {d} Clear(A) | Goal {d} On(A,B) |
| Initial {d} Clear(B) | Goal {d} On(B,C) |
| Initial {d} Clear(C) | |

There is an action called "Stack". The effect of the stack action is $On(x, y)$: block $x$ is on top of block $y$. For the action to be successfully executed, the conditions $Clear(x)$ and $Clear(y)$ must hold: neither block $x$ or block $y$ have a block on them. Planning introduces two stacking actions and the following temporal constraints.

| *Stacking Action* | *Stacking Action* |
| --- | --- |
| Stack(A,B) {bi,mi} Initial | Stack(B,C) {bi,mi} Initial |
| Stack(A,B) {d} Clear(A) | Stack(B,C) {d} Clear(B) |
| Stack(A,B) {f} Clear(B) | Stack(B,C) {f} Clear(C) |
| Stack(A,B) {m} On(A,B) | Stack(B,C) {m} On(B,C) |

A graphical representation of the IA network for this planning problem is shown in Figure 2a. Two fundamental tasks are determining whether the temporal information is consistent, and, if so, finding one or more scenarios that are consistent with the temporal information. An IA network is *consistent* if and only if there exists a mapping $M$ of a real interval $M(u)$ for each event or vertex $u$ in the network such that the relations between events are satisfied (i.e., one of the disjuncts is satisfied). For example, consider the small subnetwork in Figure 2a consisting of the events On(A,B), On(B,C), and Goal. This subnetwork is consistent as demonstrated by the assignment, $M(On(A,B)) = [1, 5]$, $M(On(B,C)) = [2, 5]$, and $M(Goal) = [3, 4]$. If we were to change the subnetwork and insist that On(A,B) must be before On(B,C), no such mapping would exist and the subnetwork would be inconsistent. A *consistent scenario* of an IA network is a non-disjunctive subnetwork (i.e., every edge is labeled with a single basic relation) that is consistent. In our planning example, finding a consistent scenario of the network corresponds to finding an ordering of the actions that will accomplish the goal of stacking the three blocks. One such consistent scenario can be reconstructed from the qualitative mapping shown in Figure 2b.

**Example 2.** Golumbic and Shamir (1993) discuss how IA networks can be used in a problem in molecular biology: examining the structure of the DNA of an organism (Benzer, 1959). The intervals in the IA network represent segments of DNA. Experiments can be performed to determine whether a pair of segments is either disjoint or intersects. Thus, the IA networks that result contain edges labeled with disjoint ({b,bi}), intersects ({m,mi,o,oi,s,si,d,di,f,fi,eq}), or $I$, the set of all basic relations—which indicates no experiment was performed. If the IA network is consistent, this is evidence for the hypothesis that DNA is linear in structure; if it is inconsistent, DNA is nonlinear (it forms loops, for example). Golumbic and Shamir (1993) show that determining consistency in this restricted version of IA networks is NP-complete. We will show that problems that arise in this application can often be solved quickly in practice.

(a) IA network for block-stacking example:

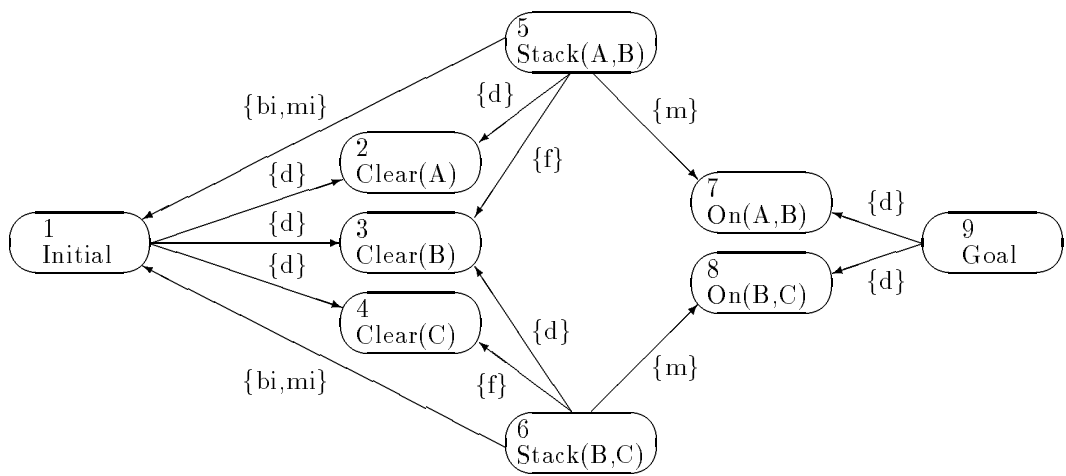

(b) Consistent scenario:

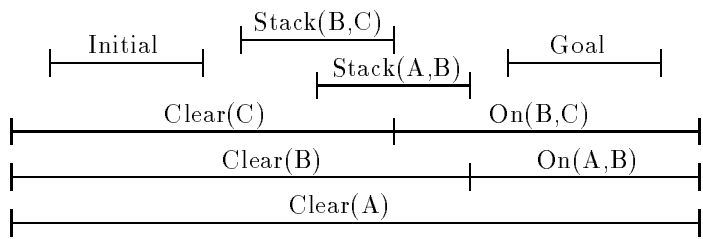

Figure 2: Representing qualitative relations between intervals

## 2.2 Test problems

We tested how well the heuristics we developed for improving path consistency and backtracking algorithms perform on a test suite of problems.

The purpose of empirically testing the algorithms is to determine the performance of the algorithms and the proposed improvements on "typical" problems. There are two approaches: (i) collect a set of "benchmark" problems that are representative of problems that arise in practice, and (ii) randomly generate problems and "investigate how algorithmic performance depends on problem characteristics ... and learn to predict how an algorithm will perform on a given problem class" (Hooker, 1994).

For IA networks, there is no existing collection of large benchmark problems that actually arise in practice—as opposed to, for example, planning in a toy domain such as the blocks world. As a start to a collection, we propose an IA network with 145 intervals that arose from a problem in molecular biology (Benzer, 1959, pp. 1614-15; see Example 2, above). The proposed benchmark problem is not strictly speaking a temporal reasoning problem as the intervals represent segments of DNA, not intervals of time. Nevertheless, it can be formulated as a temporal reasoning problem. The value is that the benchmark problem arose in a real application. We will refer to this problem as Benzer's matrix.

In addition to the benchmark problem, in this paper we use two models of a random IA network, denoted $\mathbf{B}(n)$ and $\mathbf{S}(n, p)$, to evaluate the performance of the algorithms, where $n$ is the number of intervals, and $p$ is the probability of a (non-trivial) constraint between two intervals. Model $\mathbf{B}(n)$ is intended to model the problems that arise in molecular biology (as estimated from the problem discussed in Benzer, 1959). Model $\mathbf{S}(n, p)$ allows us to study how algorithm performance depends on the important problem characteristic of sparseness of the underlying constraint graph. Both models, of course, allow us to study how algorithm performance depends on the size of the problem.

For $\mathbf{B}(n)$, the random instances are generated as follows.

**Step 1.** Generate a "solution" of size $n$ as follows. Generate $n$ real intervals by randomly generating values for the end points of the intervals. Determine the IA network by determining, for each pair of intervals, whether the two intervals either intersect or are disjoint.

**Step 2.** Change some of the constraints on edges to be the trivial constraint by setting the label to be $I$, the set of all 13 basic relations. This represents the case where no experiment was performed to determine whether a pair of DNA segments intersect or are disjoint. Constraints are changed so that the percentage of non-trivial constraints (approximately 6% are intersects and 17% are disjoint) and their distribution in the graph are similar to those in Benzer's matrix.

For $\mathbf{S}(n, p)$, the random instances are generated as follows.

**Step 1.** Generate the underlying constraint graph by indicating which of the possible $\binom{n}{2}$ edges is present. Let each edge be present with probability $p$, independently of the presence or absence of other edges.

**Step 2.** If an edge occurs in the underlying constraint graph, randomly chose a label for the edge from the set of all possible labels (excluding the empty label) where each label is chosen with equal probability. If an edge does not occur, label the edge with $I$, the set of all 13 basic relations.

**Step 3.** Generate a "solution" of size $n$ as follows. Generate $n$ real intervals by randomly generating values for the end points of the intervals. Determine the consistent scenario by determining the basic relations which are satisfied by the intervals. Finally, add the solution to the IA network generated in Steps 1–2.

Hence, only consistent IA networks are generated from $\mathbf{S}(n, p)$. If we omit Step 3, it can be shown both analytically and empirically that almost all of the different possible IA networks generated by this distribution are inconsistent and that the inconsistency is easily detected by a path consistency algorithm. To avoid this potential pitfall, we test our algorithms on consistent instances of the problem. This method appears to generate a reasonable test set for temporal reasoning algorithms with problems that range from easy to hard. It was found, for example, that instances drawn from $\mathbf{S}(n, 1/4)$ were hard problems for the backtracking algorithms to solve, whereas for values of $p$ on either side ($\mathbf{S}(n, 1/2)$ and $\mathbf{S}(n, 1/8)$) the problems were easier.

## 3. Path Consistency Algorithm

Path consistency or transitive closure algorithms (Aho, Hopcroft, & Ullman, 1974; Mackworth, 1977; Montanari, 1974) are important for temporal reasoning. Allen (1983) shows that a path consistency algorithm can be used as a heuristic test for whether an IA network is consistent (sometimes the algorithm will report that the information is consistent when really it is not). A path consistency algorithm is useful also in a backtracking search for a consistent scenario where it can be used as a preprocessing algorithm (Mackworth, 1977; Ladkin & Reinefeld, 1992) and as an algorithm that can be interleaved with the backtracking search (see the next section; Nadel, 1989; Ladkin & Reinefeld, 1992). In this section, we examine methods for speeding up a path consistency algorithm.

The idea behind the path consistency algorithm is the following. Choose any three vertices $i$, $j$, and $k$ in the network. The labels on the edges $(i, j)$ and $(j, k)$ potentially constrain the label on the edge $(i, k)$ that completes the triangle. For example, consider the three vertices Stack(A,B), On(A,B), and Goal in Figure 2a. From Stack(A,B) {m} On(A,B) and On(A,B) {di} Goal we can deduce that Stack(A,B) {b} Goal and therefore can change the label on that edge from $I$, the set of all basic relations, to the singleton set {b}. To perform this deduction, the algorithm uses the operations of set intersection ($\cap$) and composition ($\cdot$) of labels and checks whether $C_{ik} = C_{ik} \cap C_{ij} \cdot C_{jk}$, where $C_{ik}$ is the label on edge $(i, k)$. If $C_{ik}$ is updated, it may further constrain other labels, so $(i, k)$ is added to a list to be processed in turn, provided that the edge is not already on the list. The algorithm iterates until no more such changes are possible. A unary operation, inverse, is also used in the algorithm. The inverse of a label is the inverse of each of its elements (see Figure 1 for the inverses of the basic relations).

We designed and experimentally evaluated techniques for improving the efficiency of a path consistency algorithm. Our starting point was the variation on Allen's (1983) algorithm shown in Figure 3. For an implementation of the algorithm to be efficient, the intersection and composition operations on labels must be efficient (Steps 5 & 10). Intersection was made efficient by implementing the labels as bit vectors. The intersection of two labels is then simply the logical AND of two integers. Composition is harder to make efficient. Unfortunately, it is impractical to implement the composition of two labels using table lookup as the table would need to be of size $2^{13} \times 2^{13}$, there being $2^{13}$ possible labels.

We experimentally compared two practical methods for composition that have been proposed in the literature. Allen (1983) gives a method for composition which uses a table of size $13 \times 13$. The table gives the composition of the basic relations (see Allen, 1983, for the table). The composition of two labels is computed by a nested loop that forms the union of the pairwise composition of the basic relations in the labels. Hogge (1987) gives a method for composition which uses four tables of size $2^7 \times 2^7$, $2^7 \times 2^6$, $2^6 \times 2^7$, and $2^6 \times 2^6$. The composition of two labels is computed by taking the union of the results of four array references (H. Kautz independently devised a similar scheme). In our experiments, the implementations of the two methods differed only in how composition was computed. In both, the list, $L$, of edges to be processed was implemented using a first-in, first-out policy (i.e., a stack).

We also experimentally evaluated methods for reducing the number of composition operations that need to be performed. One idea we examined for improving the efficiency is

Path-Consistency$(C, n)$

1. $L \leftarrow \{(i, j) \mid 1 \leq i < j \leq n\}$
2. **while** ($L$ is not empty)
3. **do**    select and delete an $(i, j)$ from $L$
4.     **for** $k \leftarrow 1$ to $n$, $k \neq i$ and $k \neq j$
5.     **do**    $t \leftarrow C_{ik} \cap C_{ij} \cdot C_{jk}$
6.         **if** ($t \neq C_{ik}$)
7.         **then** $C_{ik} \leftarrow t$
8.             $C_{ki} \leftarrow$ Inverse$(t)$
9.             $L \leftarrow L \cup \{(i, k)\}$
10.        $t \leftarrow C_{kj} \cap C_{ki} \cdot C_{ij}$
11.        **if** ($t \neq C_{kj}$)
12.        **then** $C_{kj} \leftarrow t$
13.            $C_{jk} \leftarrow$ Inverse$(t)$
14.            $L \leftarrow L \cup \{(k, j)\}$

Figure 3: Path consistency algorithm for IA networks

to avoid the computation when it can be predicted that the result will not constrain the label on the edge that completes the triangle. Three such cases we identified are shown in Figure 4. Another idea we examined, as first suggested by Mackworth (1977, p. 113), is that the order that the edges are processed can affect the efficiency of the algorithm. The reason is the following. The same edge can appear on the list, $L$, of edges to be processed many times as it progressively gets constrained. The number of times a particular edge appears on the list can be reduced by a good ordering. For example, consider the edges $(3, 1)$ and $(3, 5)$ in Figure 2a. If we process edge $(3, 1)$ first, edge $(3, 2)$ will be updated to {o,oi,s,si,d,di,f,fi,eq} and will be added to $L$ ($k = 2$ in Steps 5–9). Now if we process edge $(3, 5)$, edge $(3, 2)$ will be updated to {o,s,d} and will be added to $L$ a second time. However, if we process edge $(3, 5)$ first, $(3, 2)$ will be immediately updated to {o,s,d} and will only be added to $L$ once.

Three heuristics we devised for ordering the edges are shown in Figure 9. The edges are assigned a heuristic value and are processed in ascending order. When a new edge is added to the list (Steps 9 & 14), the edge is inserted at the appropriate spot according to its new heuristic value. There has been little work on ordering heuristics for path consistency algorithms. Wallace and Freuder (1992) discuss ordering heuristics for arc consistency algorithms, which are closely related to path consistency algorithms. Two of their heuristics cannot be applied in our context as the heuristics assume a constraint satisfaction problem with finite domains, whereas IA networks are examples of constraint satisfaction problems with infinite domains. A third heuristic (due to B. Nudel, 1983) closely corresponds to our cardinality heuristic.

All experiments were performed on a Sun 4/25 with 12 megabytes of memory. We report timings rather than some other measure such as number of iterations as we believe this gives a more accurate picture of whether the results are of practical interest. Care was The computation, $C_{ik} \cap C_{ij} \cdot C_{jk}$, can be skipped when it is known that the result of the composition will not constrain the label on the edge $(i, k)$:

a. If either $C_{ij}$ or $C_{jk}$ is equal to $I$, the result of the composition will be $I$ and therefore will not constrain the label on the edge $(i, k)$. Thus, in Step 1 of Figure 3, edges that are labeled with $I$ are not added to the list of edges to process.

b. If the condition,

$$(\text{b} \in C_{ij} \wedge \text{bi} \in C_{jk}) \vee (\text{bi} \in C_{ij} \wedge \text{b} \in C_{jk}) \vee (\text{d} \in C_{ij} \wedge \text{di} \in C_{jk}),$$

is true, the result of composing $C_{ij}$ and $C_{jk}$ will be $I$. The condition is quickly tested using bit operations. Thus, if the above condition is true just before Step 5, Steps 5–9 can be skipped. A similar condition can be formulated and tested before Step 10.

c. If at some point in the computation of $C_{ij} \cdot C_{jk}$ it is determined that the result accumulated so far would not constrain the label $C_{ik}$, the rest of the computation can be skipped.

Figure 4: Skipping techniques

taken to always start with the same base implementation of the algorithm and only add enough code to implement the composition method, new technique, or heuristic that we were evaluating. As well, every attempt was made to implement each method or heuristic as efficiently as we could.

Given our implementations, Hogge's method for composition was found to be more efficient than Allen's method for both the benchmark problem and the random instances (see Figures 5–8). This much was not surprising. However, with the addition of the skipping techniques, the two methods became close in efficiency. The skipping techniques sometimes dramatically improved the efficiency of both methods. The ordering heuristics can improve the efficiency, although here the results were less dramatic. The cardinality heuristic and the constraintedness heuristic were also tried for ordering the edges. It was found that the cardinality heuristic was just as costly to compute as the weight heuristic but did not out perform it. The constraintedness heuristic reduced the number of iterations but proved too costly to compute. This illustrates the balance that must be struck between the effectiveness of a heuristic and the additional overhead the heuristic introduces.

For $\mathbf{S}(n, p)$, the skipping techniques and the weight ordering heuristic together can result in up to a ten-fold speedup over an already highly optimized implementation using Hogge's method for composition. The largest improvements in efficiency occur when the IA networks are sparse ($p$ is smaller). This is encouraging for it appears that the problems that arise in planning and molecular biology are also sparse. For $\mathbf{B}(n)$ and Benzer's matrix, the speedup is approximately four-fold. Perhaps most importantly, the execution times reported indicate that the path consistency algorithm, even though it is an $\mathrm{O}(n^3)$ algorithm, can be used on practical-sized problems. In Figure 8, we show how well the algorithms scale up. It can be seen that the algorithm that includes the weight ordering heuristic out performs all others. However, this algorithm requires much space and the largest problem we were able to solve was with 500 intervals. The algorithms that included only the skipping techniques were able to solve much larger problems before running out of space (up to 1500 intervals) and here the constraint was the time it took to solve the problems.

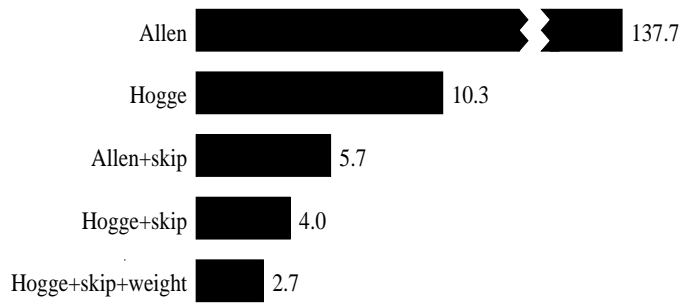

Figure 5: Effect of heuristics on time (sec.) of path consistency algorithms applied to Benzer's matrix

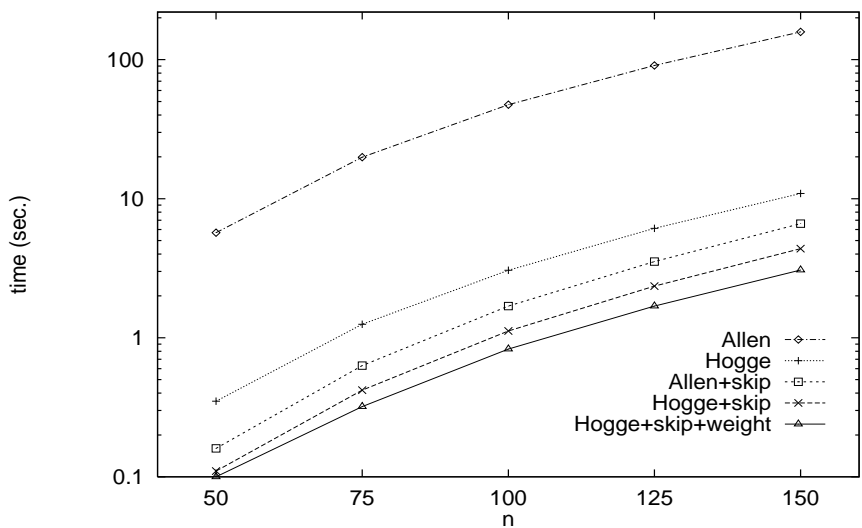

Figure 6: Effect of heuristics on average time (sec.) of path consistency algorithms. Each data point is the average of 100 tests on random instances of IA networks drawn from $\mathbf{B}(n)$; the coefficient of variation (standard deviation / average) for each set of 100 tests is bounded by 0.20

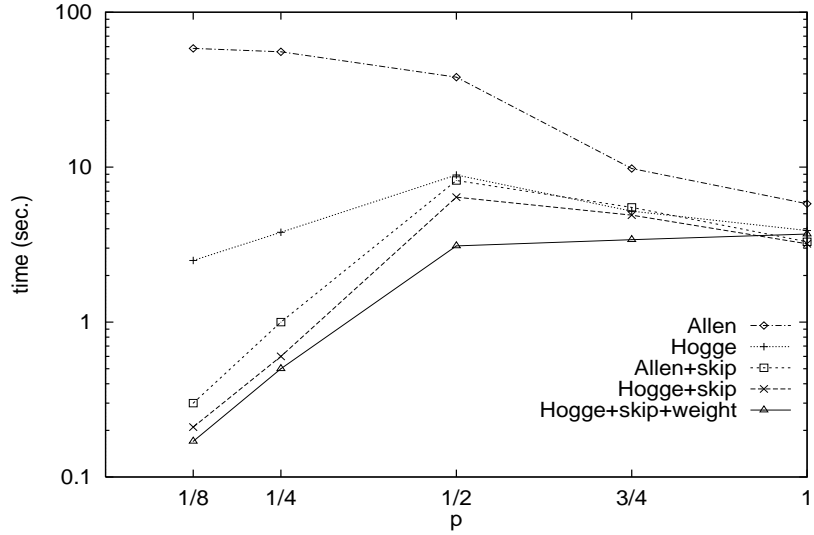

Figure 7: Effect of heuristics on average time (sec.) of path consistency algorithms. Each data point is the average of 100 tests on random instances of IA networks drawn from $\mathbf{S}(100, p)$; the coefficient of variation (standard deviation / average) for each set of 100 tests is bounded by 0.25

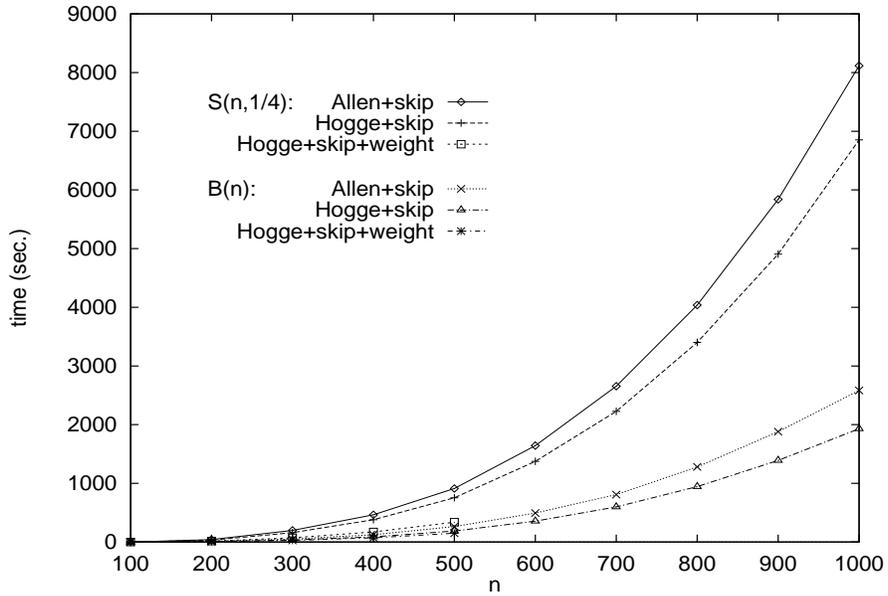

Figure 8: Effect of heuristics on average time (sec.) of path consistency algorithms. Each data point is the average of 10 tests on random instances of IA networks drawn from $\mathbf{S}(n, 1/4)$ and $\mathbf{B}(n)$; the coefficient of variation (standard deviation / average) for each set of 10 tests is bounded by 0.35

## 4. Backtracking Algorithm

Allen (1983) was the first to propose that a backtracking algorithm (Golomb & Baumert, 1965) could be used to find a consistent scenario of an IA network. In the worst case, a backtracking algorithm can take an exponential amount of time to complete. This worst case also applies here as Vilain and Kautz (1986, 1989) show that finding a consistent scenario is NP-complete for IA networks. In spite of the worst case estimate, backtracking algorithms can work well in practice. In this section, we examine methods for speeding up a backtracking algorithm for finding a consistent scenario and present results on how well the algorithm performs on different classes of problems. In particular, we compare the efficiency of the algorithm on two alternative formulations of the problem: one that has previously been proposed by others and one that we have proposed (van Beek, 1992). We also improve the efficiency of the algorithm by designing heuristics for ordering the instantiation of the variables and for ordering the values in the domains of the variables.

As our starting point, we modeled our backtracking algorithm after that of Ladkin and Reinefeld (1992) as the results of their experimentation suggests that it is very successful at finding consistent scenarios quickly. Following Ladkin and Reinefeld our algorithm has the following characteristics: preprocessing using a path consistency algorithm, static order of instantiation of the variables, chronological backtracking, and forward checking or pruning using a path consistency algorithm. In chronological backtracking, when the search reaches a dead end, the search simply backs up to the next most recently instantiated variable and tries a different instantiation. Forward checking (Haralick & Elliott, 1980) is a technique where it is determined and recorded how the instantiation of the current variable restricts the possible instantiations of future variables. This technique can be viewed as a hybrid of tree search and consistency algorithms (see Nadel, 1989; Nudel, 1983). (See Dechter, 1992, for a general survey on backtracking.)

### 4.1 Alternative formulations

Let $C$ be the matrix representation of an IA network, where $C_{ij}$ is the label on edge $(i, j)$. The traditional method for finding a consistent scenario of an IA network is to search for a subnetwork $S$ of a network $C$ such that,

(a) $S_{ij} \subseteq C_{ij}$,

(b) $|S_{ij}| = 1$, for all $i, j$, and

(c) $S$ is consistent.

To find a consistent scenario we simply search through the different possible $S$'s that satisfy conditions (a) and (b)—it is a simple matter to enumerate them—until we find one that also satisfies condition (c). Allen (1983) was the first to propose using backtracking search to search through the potential $S$'s.

Our alternative formulation is based on results for two restricted classes of IA networks, denoted here as SA networks and NB networks. In IA networks, the relation between two intervals can be any subset of $I$, the set of all thirteen basic relations. In SA networks (Vilain & Kautz, 1986), the allowed relations between two intervals are only those subsets of $I$ that can be translated, using the relations $\{<, \leq, =, >, \geq, \neq\}$, into conjunctions of relations between the endpoints of the intervals. For example, the IA network in Figure 2a is also an SA network. As a specific example, the interval relation "A {bi,mi} B" can be expressed as the conjunction of point relations, $(B^- < B^+) \wedge (A^- < A^+) \wedge (A^- \geq B^+)$, where $A^-$ and $A^+$ represent the start and end points of interval A, respectively. (See Ladkin & Maddux, 1988; van Beek & Cohen, 1990, for an enumeration of the allowed relations for SA networks.) In NB networks (Nebel & Bürckert, 1995), the allowed relations between two intervals are only those subsets of $I$ that can be translated, using the relations $\{<, \leq, =, >, \geq, \neq\}$, into conjunctions of Horn clauses that express the relations between the endpoints of the intervals. The set of NB relations is a strict superset of the SA relations.

Our alternative formulation is as follows. We describe the method in terms of SA networks, but the same method applies to NB networks. The idea is that, rather than search directly for a consistent scenario of an IA network as in previous work, we first search for something more general: a consistent SA subnetwork of the IA network. That is, we use backtrack search to find a subnetwork $S$ of a network $C$ such that,

(a) $S_{ij} \subseteq C_{ij}$,

(b) $S_{ij}$ is an allowed relation for SA networks, for all $i, j$, and

(c) $S$ is consistent.

In previous work, the search is through the alternative singleton labelings of an edge, i.e., $|S_{ij}| = 1$. The key idea in our proposal is that we decompose the labels into the largest possible sets of basic relations that are allowed for SA networks and search through these decompositions. This can considerably reduce the size of the search space. For example, suppose the label on an edge is {b,bi,m,o,oi,si}. There are six possible ways to label the edge with a singleton label: {b}, {bi}, {m}, {o}, {oi}, {si}, but only two possible ways to label the edge if we decompose the labels into the largest possible sets of basic relations that are allowed for SA networks: {b,m,o} and {bi,oi,si}. As another example, consider the network shown in Figure 2a. When searching through alternative singleton labelings, the worst case size of the search space is $C_{12} \times C_{13} \times \cdots \times C_{89} = 314$ (the edges labeled with $I$ must be included in the calculation). But when decomposing the labels into the largest possible sets of basic relations that are allowed for SA networks and searching through the decompositions, the size of the search space is 1, so no backtracking is necessary (in general, the search is, of course, not always backtrack free).

To test whether an instantiation of a variable is consistent with instantiations of past variables and with possible instantiations of future variables, we use an incremental path consistency algorithm (in Step 1 of Figure 3 instead of initializing $L$ to be all edges, it is initialized to the single edge that has changed). The result of the backtracking algorithm is a consistent SA subnetwork of the IA network, or a report that the IA network is inconsistent. After backtracking completes, a solution of the SA network can be found using a fast algorithm given by van Beek (1992).

## 4.2 Ordering heuristics

Backtracking proceeds by progressively instantiating variables. If no consistent instantiation exists for the current variable, the search backs up. The order in which the variables

**Weight.** The weight heuristic is an estimate of how much the label on an edge will restrict the labels on other edges. Restrictiveness was measured for each basic relation by successively composing the basic relation with every possible label and summing the cardinalities of the results. The results were then suitably scaled to give the table shown below.

| relation | b | bi | m | mi | o | oi | s | si | d | di | f | fi | eq |
|----------|---|----|---|----|---|----|---|----|---|----|---|----|----|
| weight   | 3 | 3  | 2 | 2  | 4 | 4  | 2 | 2  | 4 | 3  | 2 | 2  | 1  |

The weight of a label is then the sum of the weights of its elements. For example, the weight of the relation {m,o,s} is $2 + 4 + 2 = 8$.

**Cardinality.** The cardinality heuristic is a variation on the weight heuristic. Here, the weight of every basic relation is set to one.

**Constraint.** The constraintedness heuristic is an estimate of how much a *change* in a label on an edge will restrict the labels on other edges. It is determined as follows. Suppose the edge we are interested in is $(i, j)$. The constraintedness of the label on edge $(i, j)$ is the sum of the weights of the labels on the edges $(k, i)$ and $(j, k)$, $k = 1, ..., n, k \neq i, k \neq j$. The intuition comes from examining the path consistency algorithm (Figure 3) which would propagate a change in the label $C_{ij}$. We see that $C_{ij}$ will be composed with $C_{ki}$ (Step 5) and $C_{jk}$ (Step 10), $k = 1, ..., n, k \neq i, k \neq j$.

Figure 9: Ordering heuristics

---

are instantiated and the order in which the values in the domains are tried as possible instantiations can greatly affect the performance of a backtracking algorithm and various methods for ordering the variables (e.g. Bitner & Reingold, 1975; Freuder, 1982; Nudel, 1983) and ordering the values (e.g. Dechter & Pearl, 1988; Ginsberg et al., 1990; Haralick & Elliott, 1980) have been proposed.

The idea behind variable ordering heuristics is to instantiate variables first that will constrain the instantiation of the other variables the most. That is, the backtracking search attempts to solve the most highly constrained part of the network first. Three heuristics we devised for ordering the variables (edges in the IA network) are shown in Figure 9. For our alternative formulation, cardinality is redefined to count the decompositions rather than the elements of a label. The variables are put in ascending order. In our experiments the ordering is static—it is determined before the backtracking search starts and does not change as the search progresses. In this context, the cardinality heuristic is similar to a heuristic proposed by Bitner and Reingold (1975) and further studied by Purdom (1983).

The idea behind value ordering heuristics is to order the values in the domains of the variables so that the values most likely to lead to a solution are tried first. Generally, this is done by putting values first that constrain the choices for other variables the least. Here we propose a novel technique for value ordering that is based on knowledge of the structure of solutions. The idea is to first choose a small set of problems from a class of problems, and then find a consistent scenario for each instance without using value ordering. Once we have a set of solutions, we examine the solutions and determine which values in the domains

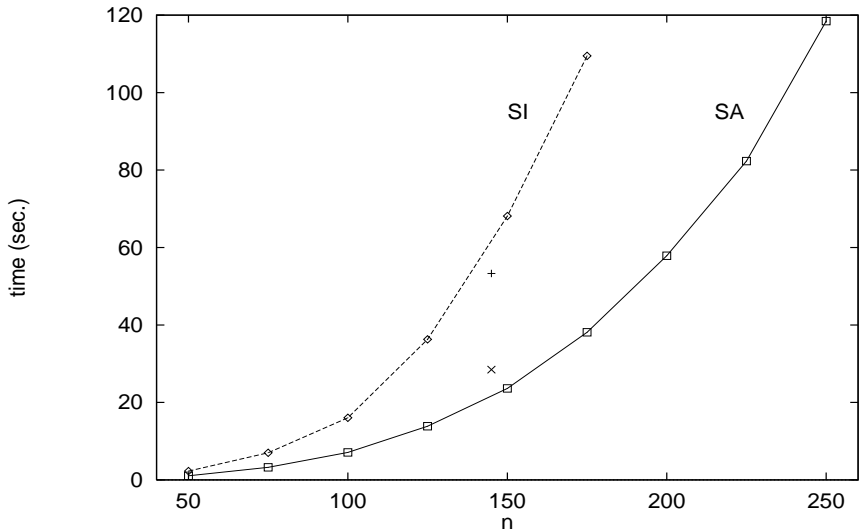

Figure 10: Effect of decomposition method on average time (sec.) of backtracking algorithm. Each data point is the average of 100 tests on random instances of IA networks drawn from **B**($n$); the coefficient of variation (standard deviation / average) for each set of 100 tests is bounded by 0.15

are most likely to appear in a solution and which values are least likely. This information is then used to order the values in subsequent searches for solutions to problems from this class of problems. For example, five problems were generated using the model **S**($100, 1/4$) and consistent scenarios were found using backtracking search and the variable ordering heuristic constraintedness/weight/cardinality. After rounding to two significant digits, the relations occurred in the solutions with the following frequency,

| relation | b, bi | d, di | o, oi | eq | m, mi | f, fi | s, si |
|----------|-------|-------|-------|-----|-------|-------|-------|
| value ($\times 10$) | 1900 | 240 | 220 | 53 | 20 | 15 | 14 |

As an example of using this information to order the values in a domain, suppose that the label on an edge is {b,bi,m,o,oi,si}. If we are decomposing the labels into singleton labels, we would order the values in the domain as follows (most preferred first): {b}, {bi}, {o}, {oi}, {m}, {si}. If we are decomposing the labels into the largest possible sets of basic relations that are allowed for SA networks, we would order the values in the domain as follows: {b,m,o}, {bi,oi,si}, since $1900 + 20 + 220 > 1900 + 220 + 14$. This technique can be used whenever something is known about the structure of solutions.

### 4.3 Experiments

All experiments were performed on a Sun 4/20 with 8 megabytes of memory.

The first set of experiments, summarized in Figure 10, examined the effect of problem formulation on the execution time of the backtracking algorithm. We implemented three versions of the algorithm that were identical except that one searched through singleton labelings (denoted hereafter and in Figure 10 as the SI method) and the other two searched through decompositions of the labels into the largest possible allowed relations for SA networks and NB networks, respectively. All of the methods solved the same set of random problems drawn from $\mathbf{B}(n)$ and were also applied to Benzer's matrix (denoted + and × in Figure 10). For each problem, the amount of time required to solve the given IA network was recorded. As mentioned earlier, each IA network was preprocessed with a path consistency algorithm before backtracking search. The timings include this preprocessing time. The experiments indicate that the speedup by using the SA decomposition method can be up to three-fold over the SI method. As well, the SA decomposition method was able to solve larger problems before running out of space ($n = 250$ versus $n = 175$). The NB decomposition method gives exactly the same result as for the SA method on these problems because of the structure of the constraints. We also tested all three methods on a set of random problems drawn from $\mathbf{S}(100, p)$, where $p = 1, 3/4, 1/2$, and $1/8$. In these experiments, the SA and NB methods were consistently twice as fast as the SI method. As well, the NB method showed no advantage over the SA method on these problems. This is surprising as the branching factor, and hence the size of the search space, is smaller for the NB method than for the SA method.

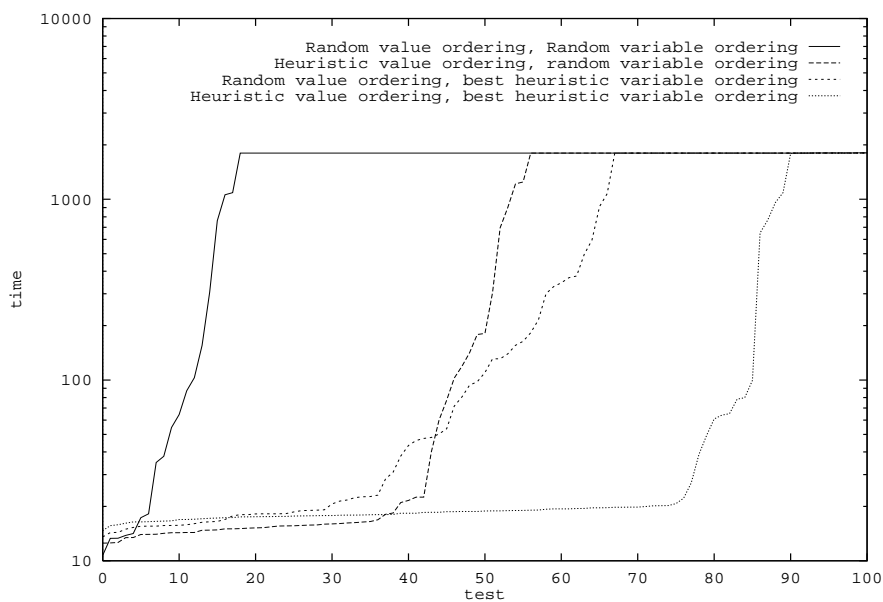

Figure 11: Effect of variable and value ordering heuristics on time (sec.) of backtracking algorithm. Each curve represents 100 tests on random instances of IA networks drawn from $\mathbf{S}(100, 1/4)$ where the tests are ordered by time taken to solve the instance. The backtracking algorithm used the SA decomposition method.

The second set of experiments, summarized in Figure 11, examined the effect on the execution time of the backtracking algorithm of heuristically ordering the variables and the values in the domains of the variables before backtracking search begins. For variable ordering, all six permutations of the cardinality, constraint, and weight heuristics were tried as the primary, secondary, and tertiary sorting keys, respectively. As a basis of comparison, the experiments included the case of no heuristics. Figure 11 shows approximate cumulative frequency curves for some of the experimental results. Thus, for example, we can read from the curve representing heuristic value ordering and best heuristic variable ordering that approximately 75% of the tests completed within 20 seconds, whereas with random value and variable ordering only approximately 5% of the tests completed within 20 seconds. We can also read from the curves the 0, 10, ..., 100 percentiles of the data sets (where the value of the median is the 50$th$ percentile or the value of the 50$th$ test). The curves are truncated at time = 1800 (1/2 hour), as the backtracking search was aborted when this time limit was exceeded.

In our experiments we found that $\mathbf{S}(100, 1/4)$ represents a particularly difficult class of problems and it was here that the different heuristics resulted in dramatically different performance, both over the no heuristic case and also between the different heuristics. With no value ordering, the best heuristic for variable ordering was the combination constraintedness/weight/cardinality where constraintedness is the primary sorting key and the remaining keys are used to break subsequent ties. Somewhat surprisingly, the best heuristic for variable ordering changes when heuristic value ordering is incorporated. Here the combination weight/constraintedness/cardinality works much better. This heuristic together with value ordering is particularly effective at "flattening out" the distribution and so allowing a much greater number of problems to be solved in a reasonable amount of time. For $\mathbf{S}(100, p)$, where $p = 1, 3/4, 1/2$, and $1/8$, the problems were much easier and all but three of the hundreds of tests completed within 20 seconds. In these problems, the heuristic used did not result in significantly different performance.

In summary, the experiments indicate that by changing the decomposition method we are able to solve larger problems before running out of space ($n = 250$ vs $n = 175$ on a machine with 8 megabytes; see Figure 10). The experiments also indicate that good heuristic orderings can be essential to being able to find a consistent scenario of an IA network in reasonable time. With a good heuristic ordering we were able to solve much larger problems before running out of time (see Figure 11). The experiments also provide additional evidence for the efficacy of Ladkin and Reinefeld's (1992, 1993) algorithm. Nevertheless, even with all of our improvements, some problems still took a considerable amount of time to solve. On consideration, this is not surprising. After all, the problem is known to be NP-complete.

## 5. Conclusions

Temporal reasoning is an essential part of tasks such as planning and scheduling. In this paper, we discussed the design and an empirical analysis of two key algorithms for a temporal reasoning system. The algorithms are a path consistency algorithm and a backtracking algorithm. The temporal reasoning system is based on Allen's (1983) interval-based framework for representing temporal information. Our emphasis was on how to make the algorithms robust and efficient in practice on problems that vary from easy to hard. For the path consistency algorithm, the bottleneck is in performing the composition operation. We developed methods for reducing the number of composition operations that need to be performed. These methods can result in almost an order of magnitude speedup over an already highly optimized implementation of the algorithm. For the backtracking algorithm, we developed variable and value ordering heuristics and showed that an alternative formulation of the problem can considerably reduce the time taken to find a solution. The techniques allow an interval-based temporal reasoning system to be applied to larger problems and to perform more efficiently in existing applications.